%% file: main.tex
\documentclass[10pt]{llncs}

\twocolumn

\usepackage{eccv}

\usepackage{eccvabbrv}

\usepackage{graphicx}
\usepackage{booktabs}

\usepackage[accsupp]{axessibility}  

\usepackage[breaklinks,colorlinks,citecolor=eccvblue]{hyperref}

\usepackage{orcidlink}

\usepackage{svg}
\usepackage{algorithm}
\usepackage{algpseudocode}
\usepackage{multirow}
\usepackage{xcolor}
\usepackage[euler]{textgreek}

\RequirePackage{etoolbox}
\RequirePackage{pgf} 
\RequirePackage{xcolor} 
\definecolor{high}{rgb}{0.3, 0.7, 0.3}
\definecolor{low}{rgb}{1.0,1.0,1.0}

\newcommand*{\opacity}{50}
\newcommand*{\spdupminval}{0.0}
\newcommand*{\spdupmaxval}{48.1}

\newcommand*{\perfminval}{-0.1}
\newcommand*{\perfmaxval}{7.4}
\newcommand{\spdupgrad}[1]{
    \ifdimcomp{#1pt}{>}{\spdupmaxval pt}{#1}{
    \ifdimcomp{#1pt}{<}{\spdupminval pt}{#1}{
         \pgfmathparse{int(round(100*(#1/(\spdupmaxval-\spdupminval))-(\spdupminval*(100/(\spdupmaxval-\spdupminval)))))}
        \xdef\tempa{\pgfmathresult}
        \cellcolor{high!\tempa!low!\opacity} #1
    }}
 }
 
\newcommand{\perfgrad}[1]{\ifdimcomp{#1pt}{>}{\perfmaxval pt}{#1}{\ifdimcomp{#1pt}{<}{\perfminval pt}{#1}{\pgfmathparse{int(round(100*(#1/(\perfmaxval-\perfminval))-(\perfminval*(100/(\perfmaxval-\perfminval)))))}\xdef\tempa{\pgfmathresult}\cellcolor{high!\tempa!low!\opacity}#1}}}

\usepackage{graphicx}
\usepackage{amsmath}
\usepackage[hang,flushmargin]{footmisc}

\usepackage{todonotes} 

\newcommand{\distas}[1]{\mathbin{\overset{#1}{\kern\z@\sim}}}%

\usepackage{colortbl}
\definecolor{ourbg}{rgb}{0.9, 0.9, 1.}  

\input{math_commands.tex}

\newcommand{\ourmethod}{\textit{ClassAct }}

\usepackage{makecell}

\title{Bad Students Make Great Teachers: \\ Active Learning Accelerates Large-Scale Visual Understanding}

\titlerunning{Bad Students Make Great Teachers}

\author{
\begin{tabular}{ccc}
Talfan Evans$^{1,*}$ &
Shreya Pathak$^{1,*}$ &
Hamza Merzic$^{1,2,*}$ \\
Jonathan Schwarz$^{1,\dagger}$ &
Ryutaro Tanno$^{1}$ &
Olivier J. Hénaff$^{1,*}$ \\
\end{tabular}
\vspace{0.5em} \\
}

\authorrunning{T. Evans et al.}

\institute{$^{1}$Google DeepMind \quad $^{2}$University College London}

\begin{document}
\maketitle

\def\thefootnote{*}\footnotetext{Equal technical contribution. Email correspondence to $<$talfan@deepmind.com$>$ and $<$henaff@deepmind.com$>$. $^{\dagger}$Current affiliation: Harvard University, work done while at Google DeepMind.}

\input{sec/0_abstract}

\input{sec/1_intro}
\input{sec/2_related_work}

\input{sec/3_methods}

\input{sec/4_results}

\input{sec/5_discussion}


%
%
\bibliographystyle{splncs04}
\bibliography{main}

\input{sec/6_appendix}

\end{document}

%% file: math_commands.tex
\usepackage{amsmath,amsfonts,bm}

\def\1{\bm{1}}

\def\vx{{\bm{x}}}

\def\vz{{\bm{z}}}

\DeclareMathAlphabet{\mathsfit}{\encodingdefault}{\sfdefault}{m}{sl}
\SetMathAlphabet{\mathsfit}{bold}{\encodingdefault}{\sfdefault}{bx}{n}

\newcommand{\x}{$\times$}


%% file: sec/0_abstract.tex
\section*{\centering Abstract}
\textit{
Power-law scaling indicates that large-scale training with uniform sampling is prohibitively slow. 
Active learning methods aim to increase data efficiency by prioritizing learning on the most relevant examples. 
Despite their appeal, these methods have yet to be widely adopted since no one algorithm has been shown to a) generalize across models and tasks b) scale to large datasets and c) yield overall FLOP savings when accounting for the overhead of data selection.
In this work we propose a method which satisfies these three properties,
leveraging small, cheap proxy models to estimate ``learnability'' scores for datapoints, which are used to prioritize data for training much larger models.
As a result, models trained using our methods -- \textit{ClassAct} and \textit{ActiveCLIP} -- require 46\% and 51\% fewer training updates and up to 25\% less total computation to reach the same performance as uniformly-trained visual classifiers on JFT and multimodal models on ALIGN, respectively. 
Finally, we find our data-prioritization scheme to be complementary with recent data-curation and learning objectives, yielding a new state-of-the-art in several multimodal transfer tasks.
}

%% file: sec/1_intro.tex
\section{Introduction}
\label{sec:intro}

Power-law scaling for vision and language models \cite{kaplan2020scaling, zhai2022scaling} indicates that incremental improvements in model performance require orders of magnitude increases in computation.
One of the key features of these empirical power-laws is that training data is sampled uniformly. In contrast, active data selection 
prioritizes 
computation 
on the data that maximally contributes to task performance \cite{lindley1956measure,mackay1992information,settles2009active}, with the ultimate goals of improving data efficiency and reducing the cost of training.  
However, active data selection has yet to become a mainstay of large model training, since no existing algorithm satisfies a) robustness to the choice of model and training task, b) scalability to large datasets and architectures, and c) favourable end-to-end compute efficiency vs. training with uniform samples. 

\begin{figure}[t]
  \centering
  \includegraphics[width=1.0\linewidth]{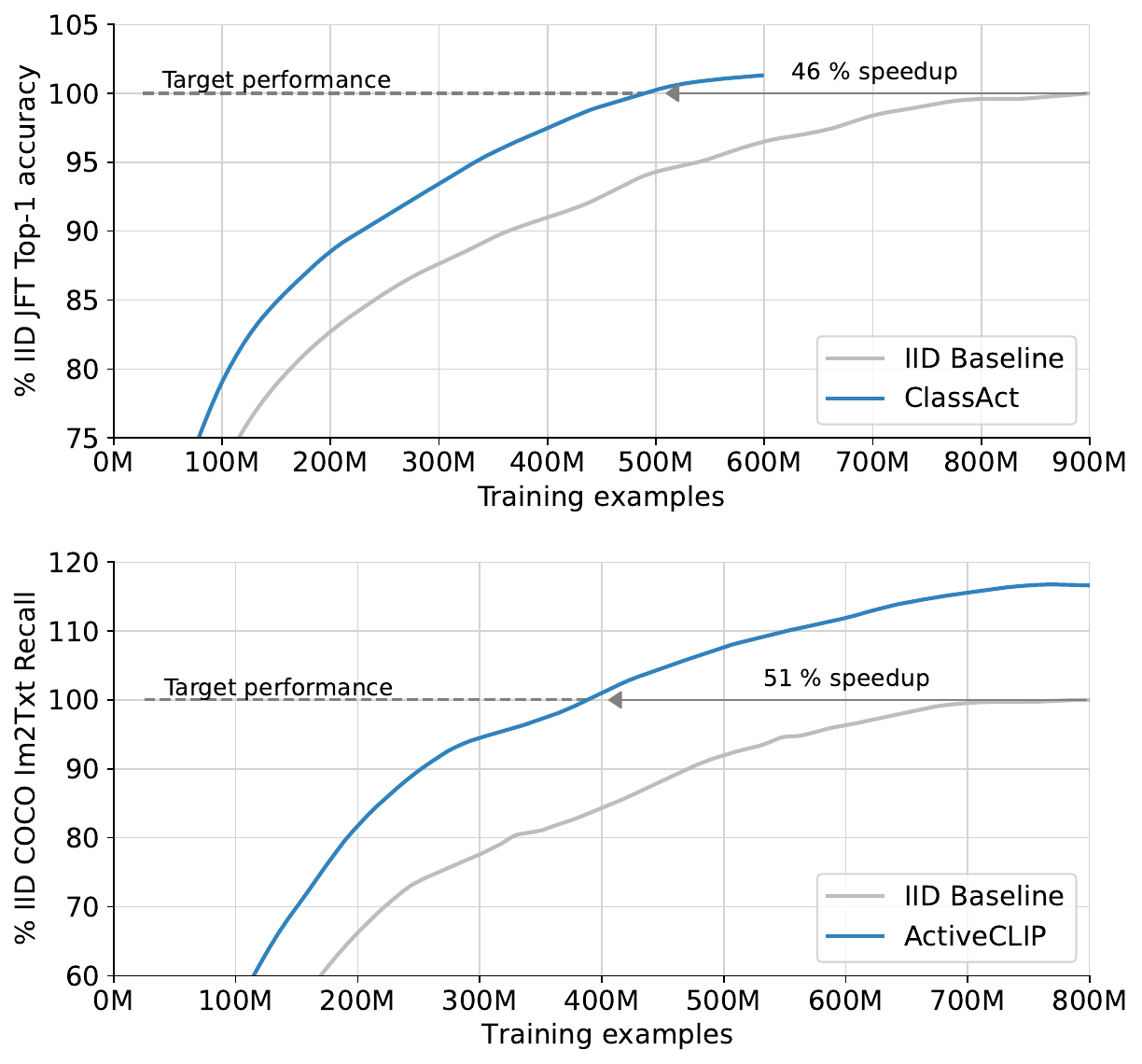}
  \caption{\textbf{Active learning accelerates large-scale visual understanding}. For large-scale classification and multimodal learning tasks, prioritised training on data selected using our active selection methods \textit{ClassAct} (\textbf{left}) and \textit{ActiveCLIP} (\textbf{right}) requires significantly fewer updates to reach the final performance of uniform training. \vspace{-2em}}  
  \label{fig:fig_0_teaser}
\end{figure}

In the first instance, data selection based on hand-engineered filters (e.g. removing incorrectly shaped images or that only contain a single colour \cite{alayrac2022flamingo}) can trivially improve training efficiency at minimal computational overhead. However, such heuristics are limited in their effectiveness by the expertise of the human designer, and are not guaranteed to transfer to training of different models, data modalities, or tasks and incur significant effort to develop and tune.

In contrast, model-based curation methods, which use the loss of the model itself to score examples, have shown promise 
by focusing training on `hard' and omitting `easy' data \cite{sorscher2022beyond}, but also the opposite (to exclude noise or other low-quality data) in both language modeling \cite{gunasekar2023textbooks} and multimodal learning \cite{hessel2021clipscore}. 
However, these methods often spend as much computation on the curation of datasets as is gained from subsequent pretraining, making them less compute-efficient than training on uniformly-sampled data. Finally, while several compute-efficient methods have been successfully deployed at small scale \cite{coleman2019selection}, they generally do not scale to even medium-sized datasets such as ImageNet. 


In this work, we propose an algorithm that satisfies the three properties of generality, scalability, and \textit{compute-positivity}. The proposed framework uses small proxy models to compute \textit{learnability} scores for candidate training data, resulting in significant training efficiency gains at an almost negligible overhead over standard uniform-training. We test two instantiations of this framework,  \textit{ClassAct} and \textit{ActiveCLIP} / \textit{ActiveSigLIP}, 
for large scale classification and multimodal pretraining, respectively. Our findings are summarized below:




\vspace{0.5em}\noindent
\textbf{Benchmarking heuristics for large-scale pretraining:} We investigate loss- and learnability-based prioritization \cite{hessel2021clipscore,sorscher2022beyond,mindermann2022prioritized} for large-scale classification and find that pretrained reference models are an essential component for accelerating learning, producing efficiency gains of up to 46\%.

\vspace{0.5em}\noindent
\textbf{Generalizing selection policies across model scale:} Secondly, we show that smaller models act as effective proxies for much larger ($\sim$1000\x) models in the context of learnability but \textit{not} loss-based scoring, resulting in \textit{compute-positive} gains of up to 25\% over uniform training, a first in the context of large-scale pretraining.

\vspace{0.5em}\noindent
\textbf{Accelerating multimodal pretraining:} 
Using reference models pretrained on small, clean datasets, we substantially accelerate pretraining on much larger, noisier datasets. Moreover, we find ActiveCLIP to be complementary to recent data-curation techniques \cite{gadre2023datacomp} and learning objectives \cite{zhai2023sigmoid}, yielding a new state-of-the-art in several multimodal understanding tasks.

\vspace{0.5em}\noindent
\textbf{Amortizing data selection policies:} Data-selection policies trained on one task can also accelerate the training of subsequent models on different but related tasks, suggesting that such policies can be easily derived from pre-trained models.

\vspace{0.5em}\noindent
\textbf{Simplifying active-learning with online reference models:} Lastly, we demonstrate that pre-trained reference models may not be necessary at all, where these models are small and can be trained in parallel on larger batches than the learner model, while remaining \textit{compute-positive}.




%% file: sec/2_related_work.tex
\section{Related Work}
\label{sec:rw}


\textbf{Data pruning.} One approach to data-selection is to identify and sub-select data ahead of training. For example, \cite{paul2021deep} and \cite{sorscher2022beyond} show that the training loss and gradients can be used to discard large portions of small-to-medium sized datasets (\eg CIFAR10 and ImageNet) with little loss in performance.  
These methods have since been deployed for the curation of web-scale datasets in both language modeling \cite{marion2023less} and multimodal learning \cite{gadre2023datacomp, abbas2023semdedup, mahmoud2023sieve}, demonstrating large reductions in the amount of data required together with performance improvements. 
However, in the single-epoch regime that is becoming typical of large model training \cite{hoffmann2022training,kaplan2020scaling}, pre-filtering can be as expensive as learning from it, a shortcoming which we address in this work. 
Nevertheless, we show that our method for dynamic data selection is complementary to and benefits from such data curation techniques. 

\vspace{0.5em}\noindent
\textbf{Online active learning.} Unlike data-pruning, online active learning continuously filters data throughout training and applies naturally to the semi-infinite, single-epoch regime.
Online Batch Selection \cite{loshchilov2015online} 
scores and filters using the learner model, which has the theoretical advantage that the importance of data can be determined relative to the current state of the learner. 
In terms of metrics, the Reducible Holdout Loss (RHO) \cite{mindermann2022prioritized} also uses the concept of a reference model to identify learnable data points not yet well represented. 
Other proposed heuristics include memorization for long-tailed data \cite{feldman2020does} and assigning ``complexity'' scores based on the number of times the example is forgotten during training \cite{toneva2018empirical}. None of these approaches however have demonstrated that the cost of scoring can be reduced to the point of justifying learner efficiency gains. 

\vspace{0.5em}\noindent
\textbf{Compute-efficient data selection.} Several works have demonstrated the benefits of selecting data based on simple heuristics, such as low-level image properties \cite{alayrac2022flamingo} or proximity to high-quality text corpora \cite{prabhu2019sampling,brown2020language, chowdhery2023palm,xie2023data}. While cheap to compute, these statistics often require domain-specific knowledge which limits their applicability across tasks. Domain-agnostic methods such as core-sets alleviate this by selecting data based on the geometry of their embeddings \cite{har2005smaller,campbell2018bayesian} which can be efficiently computed, however these algorithms generally do not scale to large-scale datasets \cite{coleman2019selection}.   Most related to our work is DoReMi \cite{xie2023doremi} which uses domain-general, scalable, and compute-efficient proxy models for the simpler problem of determining optimal data-mixtures for the subsequent training of a larger language model.

\vspace{3em}

%% file: sec/3_methods.tex
\section{Methods}
\label{sec:method}



\subsection{Data selection as prioritized replay}
\label{sec:meth-framework}



We use online batch selection \cite{loshchilov2015online} to apply our scoring heuristics 
to standard visual learning tasks: firstly, we sample uniformly from the training set $\vx_i \stackrel{\mathcal{U}}{\sim} \mathcal{D}$ and compute a score $s_i = s(\vx_i | \theta) \in \mathbb{R}$ to each data point $\vx_i$ using model parameters $\theta$. Given a large enough collection of scored examples stored in a memory bank $\mathcal{M} = \{\vx_i\}_{i \in (0, \dots, M-1)}$, we then sample \textit{non-uniformly} according to their scores $\vx_i \stackrel{\pi}{\sim} \mathcal{M}$ \cite{schaul2015prioritized}, where 
$\pi(\vx_i) =  \text{Softmax}(\{s_i\}_{i \in (0, ..., M-1) } )$. A batch of such examples is used to update the learner model. Following convention in reinforcement learning, we refer to the scoring and target models as \textit{actors} and \textit{learners} respectively.

%




\subsection{Statistics for data selection}
\label{sec:meth-heuristics}

We explore a few statistics for model-based prioritization, grouped into two categories. 

\vspace{0.5em}\noindent
\textbf{Example difficulty:} given the current state of the learner, an intuitive prioritization scheme might favour `difficult' examples (as measured by their training loss), while removing `easy' examples that are trivially classified and which yield small gradients. This loss-based prioritization:
\begin{equation}
\label{eq:hard}
s^\textrm{hard}(\vx_i | \theta)  =  \ell(\vx_i | \theta)    
\end{equation}
can use the current parameters of the learner $\theta^t$ or those of a fixed model $\theta^*$. The opposite argument can been made for favoring examples that are \textit{easily} solved by a well-trained model, as such a prioritization removes the noisy examples present in large-scale datasets: 
\begin{equation}
\label{eq:easy}
s^\textrm{easy}(\vx_i | \theta)  =  - \ell(\vx_i | \theta)
\end{equation}
This 
scheme is commonly used in multimodal learning for identifying high-quality examples with pre-trained models \cite{hessel2021clipscore, schuhmann2021laion, schuhmann2022laion}.

\vspace{0.5em} \noindent
\textbf{Example \textit{learnability}:} Given that favoring easy and hard examples target different and potentially orthogonal properties of the data, a natural question is whether these policies can be combined. \textit{Learnability} criteria straightforwardly combine the two as
\begin{align}
\label{eq:learn}
s^\textrm{learn}(\vx_i | \theta^t, \theta^*)
& =  s^\textrm{hard}(\vx_i | \theta^t) + s^\textrm{easy}(\vx_i | \theta^*) \\
& =  \ell(\vx_i | \theta^t) - \ell(\vx_i | \theta^*),
\end{align}
favoring examples that are easily solved by a well-trained model $\theta^*$ but challenging to the learner in its current state $\theta^t$, such that more computation dedicated to this example could lower its loss. Conversely, examples that are trivially classified by the learner (or mislabeled) will yield low (or high) losses for \textit{both} the current learner and the well-trained one, leading to low learnability scores. 

A special case of learnability scores (the \textit{RHO} loss, \cite{mindermann2022prioritized}) uses a model $\theta^\textrm{ho}$ specifically trained on a held-out dataset to ensure the independence of its predictions from those of the current learner $s^\textrm{learn}(\vx_i | \theta^t, \theta^\textrm{ho})$. We assess in Section \ref{sec_policies_scale} whether this is necessary when training on large-scale image datasets.




\begin{figure*}
    \centering
    \includegraphics[width=1.0\linewidth]{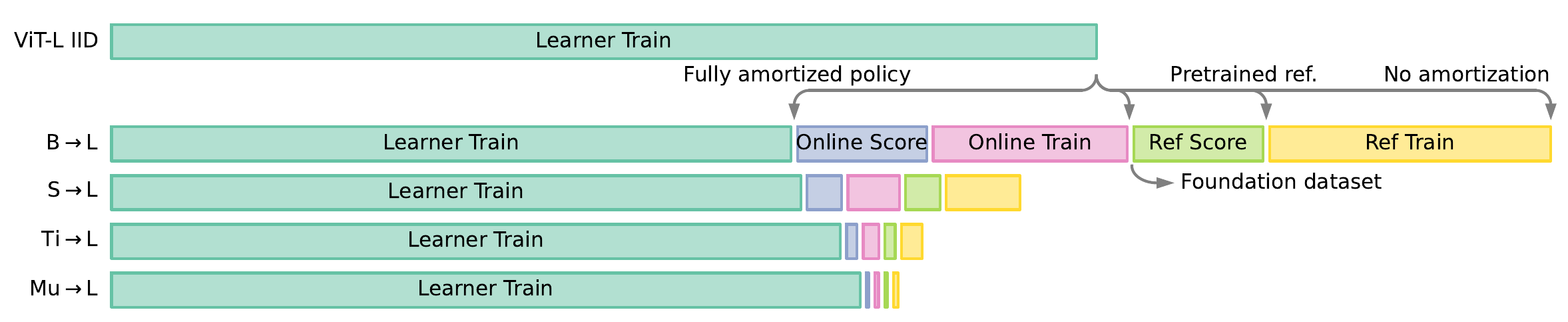}
    \caption{\textbf{Amortizing the cost of data selection}. Drawn to scale: length of bars indicates number of FLOPs required to reach the accuracy of a ViT-L trained with uniform sampling (“ViT-L Uniform Sampling”, see Figure \ref{fig:fig_3_generalisation}). Expensive model policies (e.g. a ViT-B scores data for the ViT-L learner, or `B $\rightarrow$ L') produce large learner speedups, at the expense of the additional FLOPs associate with data selection. This overhead can be reduced by deriving the data-selection policies from smaller models (e.g. ViT-S, ViT-Ti or ViT-Mu score data for the ViT-L learner), at the expense of marginal decreases in the learner speedup. Costs could be additionally amortized by using off-the-shelf reference models, removing the need to train from scratch (yellow). Since the reference model is fixed throughout training, scores can be assigned once to a `foundation dataset' and amortized across many training runs (lime green; \cite{sorscher2022beyond, mindermann2022prioritized}). Since the online model is independent of the learner model and generalizes across scale, data selection policies can also be distilled as a fixed ordering of a given dataset (a `foundation curriculum').\vspace{-1em}}
    \label{fig:bar_plot_amortization}
\end{figure*}

\subsection{Unlocking compute-positive training}
\label{sec:meth-compute}

Scoring requires inference passes over the \textit{actor} and \textit{learner} models. We assume that gradient updates to cost 3x inference passes. The cost of scoring data $F$ thus scales with the proportion of data which is being rejected (\eg retaining only 20\% of the data requires 5 inference passes per trained batch). The requirements for compute-positivity can therefore be expressed as:
\begin{equation}
\underbrace{
\Big( 3 F_{\text{learn}} + \rho F_{\text{act}} \Big) \beta + 3 F_{\text{ref}}
}_{\text{Active Learning}}
<
\underbrace{\vphantom{\frac{1}{1}} 3 F_{\text{learn}}}_{\text{Uniform Sampling}}
\end{equation}
where $F_{\text{act}}$ is the cost of scoring an example, $\rho$ is the number of examples scored per training example, and $\beta$ is the efficiency gain (in terms of the learner update) relative to training with uniform sampling 
(see Appendix Section \ref{app:flops} for more details). 
The RHS term is the cost per update of uniform sampling. The first LHS term inside the brackets is the cost of the learner training during AL, the second term is the scoring cost and the last term outside the brackets the cost of training the reference model. We illustrate in Figure \ref{fig:bar_plot_amortization} the different contexts in which parts of this computation may be effectively amortized.

In the large-scale training regime where data is neither repeated nor seen before, 
compute-positivity requires that either or all of the reference model training, actor scoring and learner efficiency $\beta$ terms must be made smaller to produce net savings relative to uniform sampling. 
Typical prioritization schemes can produce saving on the order of 50\% (ie $\beta = 0.5$), suggesting that savings must also be made by down-scaling the other terms $F_{\text{act}}$ and $F_{\text{ref}}$.

\vspace{0.5em}\noindent
\textbf{Cost of easy-reference scoring.} While both the cost of the reference model and example scoring can be scaled down in the case of easy-reference scoring with small models ($F_{\text{act}} = F_{\text{ref}}$, see Equation \ref{eq:easy}), it is unclear whether the efficiency gains $\beta$ are robust to this down-scaling (see section \ref{sec_policies_scale}).

\vspace{0.5em}\noindent
\textbf{Cost of RHO \textit{learnability} scoring.} The original definition of learnability scores \cite{mindermann2022prioritized} requires inference passes through both the learner and a reference model ($F_{\text{act}} = F_{\text{ref}} + F_{\text{learn}}$, see Equation \ref{eq:learn}), meaning that although the  cost of the reference model can be reduce by using a smaller model, the cost of example scoring cannot. 

\vspace{0.5em}\noindent
\underline{\textbf{Cost of ClassAct / ActiveCLIP.}} For this reason, we explore whether replacing the learner model in term 1 of Equation \ref{eq:learn} with a much smaller model can still produce comparable learner efficiency gains to those already observed (see Appendix Algorithm 1). Specifically, we introduce a third ``online'' model, which has the same architecture and size as the reference model, but is trained in parallel with the learner. In this case, the cost of scoring examples reduces to:

\begin{equation}
F_{\text{act}} = F_{\text{ref}} + F_{\text{online}} = 2 F_{\text{ref}}   
\end{equation}
and can be scaled down along with the reference model. 
We instantiate our method for two canonical pre-training tasks: visual classification and multimodal learning, which we call ClassAct, ActiveCLIP and ActiveSigLIP respectively. In Appendix Section \ref{sec:distributed_setup}, we describe an asynchronous active learning framework where scoring and learning is performed in parallel on separate devices. This setup does not affect the FLOP calculations but can mitigate the overhead in time even when using non-approximate scoring.

\begin{algorithm}
\small
\caption{ClassAct / ActiveCLIP}
\begin{algorithmic}[1]
\State \textbf{Input:} Randomly initialized learner model $\theta_l$ and small online model $\theta_o$, small pre-trained reference model $\theta_r$. 
Models use loss $\ell_\textrm{act}$ for scoring data and $\ell_\textrm{learn}$ for computing updates. Dataset $\mathcal{D}$, batch size $B$, sub-batch size $b<B$.
\While{training}
    \State $X \sim \mathcal{D}$, where $|X| = B$ \Comment{Sample uniformly}
    \State $S = \ell_{\text{act}}(X | \theta_o) - \ell_{\text{act}}(X | \theta_r)$ \Comment{Get scores}
    \State $I \sim \text{SoftMax}(S)$, where $|I| = b$ \Comment{Sample indices}
    \State $Y = X[I]$ \Comment{Collect sub-batch}
    \State $\theta_l \leftarrow \text{Adam}[\nabla_{\theta_l} \ell_{\text{learn}}(Y|\theta_l)]$ \Comment{Update learner model}
    \State $\theta_o \leftarrow \text{Adam}[\nabla_{\theta_o} \ell_{\text{learn}}(Y|\theta_o)]$ \Comment{Update online model}
\EndWhile
\end{algorithmic}
\label{alg:classact_algo}
\end{algorithm}

\subsection{Losses for canonical visual pre-training tasks}
\label{sec:meth-losses}

For visual classification with ClassACT, we use the standard cross-entropy loss for both actors and learners.
For multimodal learning with ActiveCLIP, learners optimize the contrastive loss $\ell_\textrm{learn} = \ell^\textrm{im,txt}_\textrm{learn} + \ell^\textrm{txt,im}_\textrm{learn}$, whereas the actor loss $\ell_\textrm{act}$ is simply the dot-product similarity between image and text embeddings:
\begin{align}
\label{eq:con}
\ell_\textrm{act}(\vx_i | \theta) & = - \vz^\textrm{im}_i {\cdot} \vz^\textrm{txt}_i \\ \ell^\textrm{im,txt}_\textrm{learn}(\vx_i | \theta) & = - \log \frac{\exp( \vz^\textrm{im}_i {\cdot} \vz^\textrm{txt}_i)}{\sum_j \exp(    \vz^\textrm{im}_i {\cdot} \vz^\textrm{txt}_j )}
\end{align}
where $\vz^\textrm{im}_i = f^\textrm{im}(\vx_i ; \theta)$ and $\vz^\textrm{txt}_i = f^\textrm{txt}(\vx_i ; \theta)$ are image and text embeddings respectively, and $\ell^\textrm{txt,im}_\textrm{learn}(\vx_i ; \theta)$ is defined analogously to $\ell^\textrm{im,txt}_\textrm{learn}(\vx_i ; \theta)$. Similarly, ActiveSigLIP instead uses the sigmoid loss \cite{zhai2023sigmoid} for the learner's objective and $\ell_\textrm{act}$ for scoring.


%% file: sec/4_results.tex
\section{Experiments}
\label{sec:exp}

All our experiments were conducted with Vision Transformers \cite{dosovitskiy2020image} for which strong baselines are available across model sizes \cite{zhai2022scaling}. Unless specified, we adopt models with patch-size 16 throughout (ViT-S refers to ViT-S/16 and similar). We consider two canonical tasks for large-scale pretraining: classification on JFT-300M \cite{sun2017revisiting} and multimodal contrastive learning  \cite{radford2021learning} on large image-text datasets. When pre-training with JFT classification we use held-out classification performance as the evaluation metric. When pre-training on image-text data we evaluate with standard multimodal transfer tasks: ImageNet zero-shot classification and image-to-text / text-to-image retrieval on COCO.

\begin{figure}
    \centering
    \includegraphics[width=1.0\linewidth]{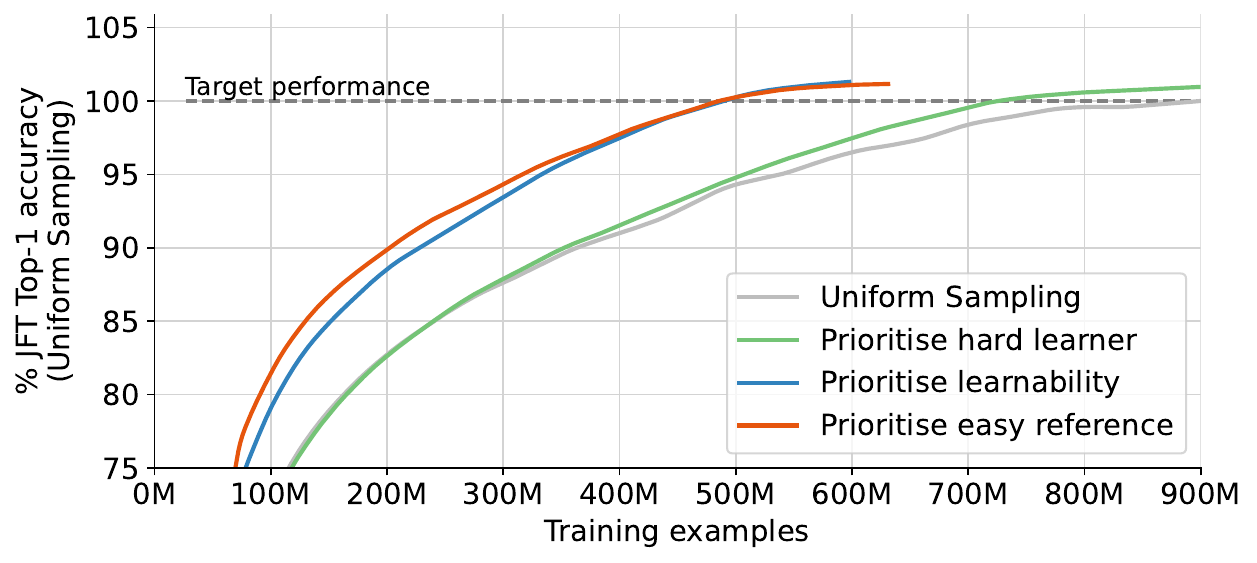}
    \caption{\textbf{Evaluation of loss-based data-selection criteria for large-scale classification}. We train a ViT-B on JFT-300M with different data-selection policies. Prioritising hard data under the learner (green curve) produced marginal gains over the uniform sampling baseline. Prioritizing data using both \textit{learnability} (blue curve, \cite{mindermann2022prioritized}) and \textit{easy reference} prioritization (red curve, \cite{hessel2021clipscore}) produced significant speedups and performance gains. \vspace{-1em}}
    \label{fig:fig_2_heuristics}
\end{figure}

\begin{figure*}
    \centering
    \includegraphics[width=\textwidth]{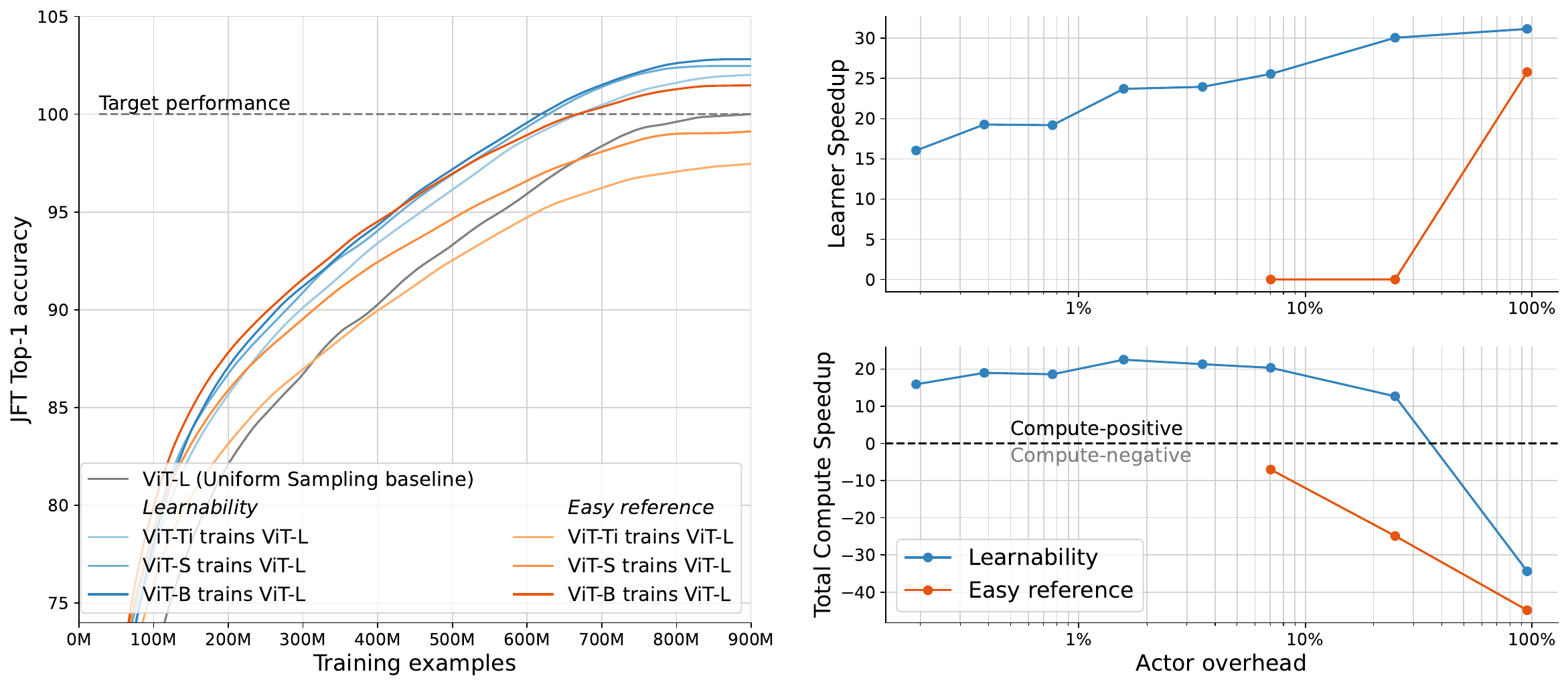}
    \caption{\textbf{Generalization of data-selection policies across models scales.} \textbf{Left:} We train a ViT-L for 3 epochs on JFT using uniform sampling (grey) or prioritized data sampling using example learnability (blue) or low-loss under the reference model (red). Example scores are computed using ViT-B actors (dark), or cheaper ViT-S or ViT-Tiny models (light). While both example learnability and ``easy reference'' yield good speedups with expensive actors, learnability criteria are much more robust to approximate scoring. \textbf{Top right:} Learner (ViT-L) speedup is computed as the fraction of learner iterations saved in order to attain the baseline's top performance. Actor overhead is computed as the additional computation in FLOPs required to score examples with a particular actor architecture (varying from ViT-Mu to ViT-L, see Appendix Table 
    \ref{tab:small_vits}). Example learnability yields robust learner speedups across actor scales, ``easy reference'' scoring does not. \textbf{Lower right:} total compute efficiency is calculated as a product of learner efficiency and actor overhead, indicating the amount of computation required to reach baseline performance. Approximate actors (\ie ViT-S or smaller) computing example learnability enable total compute speedups, other schemes do not. \vspace{-1em}
    }
    \label{fig:fig_3_generalisation}
\end{figure*}

Throughout, we will refer to the large batch of size $B$ sampled uniformly from the training data as the `super-batch', and the prioritised smaller batch of size $b<B$ as the `sub-batch'. In all our experiment, we filter 50\% of uniformly sampled data such that $\rho = B/b = 2$, although more aggressive filtering regimes warrant investigation \cite{sorscher2022beyond}. 


\subsection{Evaluating loss-based scoring heuristics in the large-data regime}

We begin by evaluating loss- and learnability-based heuristics on their ability to accelerate supervised classification on JFT (Fig. \ref{fig:fig_2_heuristics}). Arguably the most intuitive method to score data is to prioritise training on data with high loss under the learner (\textit{hard learner}). In our experiments, this strategy (Eq. \ref{eq:hard}) only marginally improved performance over the uniform sampling baseline, despite requiring an additional inference pass over the super-batch. This is perhaps not surprising - data points with high loss may also be unlearnable due to \eg label noise, such that training on those data points does not result in the model performing any better on the held-out test set. Large scale datasets are more likely to be noisy.

Scoring methods based on pre-trained reference models performed much better---both \textit{easy reference} (equation \ref{eq:easy}) and \textit{learnability} (equation \ref{eq:learn}) -based prioritization produced significant gains over uniform sampling. Here, we pre-trained an identical ViT-B for the same 3 epochs to use as a reference model for a second training run displayed above.
producing speed-ups of ~33\% (Fig. \ref{fig:fig_2_heuristics}).

\begin{figure*}
    \centering
    \includegraphics[width=1.0\linewidth]{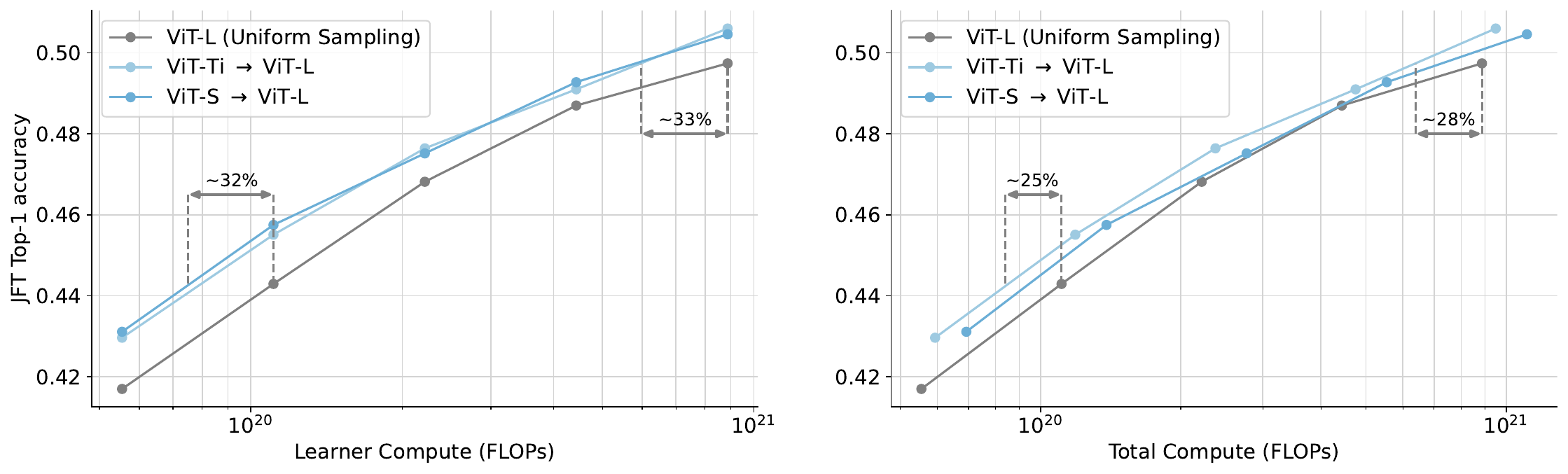}
    \caption{\textbf{Scaling laws for active learning}. We trained a baseline ViT-L over a range of compute budgets (for which ViT-L is compute optimal, see Zhai et al., 2021). We also trained the same ViT-L with both ViT-Ti and ViT-S reference policies, pre-trained for the same number of epochs. \textbf{Left}: Small model policies produce robust savings in learner compute. \textbf{Right}: When accounting for total compute (learner + actor training and data scoring), small model policies in all compute budgets produce FLOP savings over training with uniform samples. These scaling laws generalize those measured empirically in the uniform sampling setting \cite{zhai2022scaling} to the case of non-uniform data selection.
    }
    \label{fig:fig_1_scaling}
\end{figure*}

\input{tables/table1}

\subsection{Generalising data-selection policies across scale}\label{sec_policies_scale}
The speed-ups afforded by \textit{learnability} prioritization (Fig. \ref{fig:fig_2_heuristics}) come at the cost of the additional inference passes required to score the data during learner training, plus the cost of training the reference model. This makes the overall gains strongly \textit{compute-negative} relative to training with uniform samples. Even if the size of the reference model is scaled down \cite{mindermann2022prioritized}, these methods still incur the cost of additional learner inference passes to score the data during training. 




\textbf{Unlocking compute-positive active learning.} To address this issue, we introduce a set of down-scaled models with the same ViT architecture that we use to score data for training a larger ViT-L model (see Methods Section \ref{sec:meth-compute}. We use ViT-B, ViT-S or ViT-Ti variants (which are ~4\x, 13\x \ and 47\x \ cheaper than the learner) for both the online and reference model 
(see Appendix Algorithm 1). 
In Figure \ref{fig:fig_3_generalisation} (left) we assess the impact of these cheaper scoring models on learner efficiency. First, we find that easy reference prioritization to be very sensitive to the capacity of the scoring model: while ViT-B scoring models yield reasonable gains over uniform sampling, prioritizing with ViT-S and ViT-Ti scoring models underperforms significantly (Fig. \ref{fig:fig_3_generalisation}, red curves). 

In contrast, we find that \textit{learnability} based prioritization yields robust gains, even when the scoring models are significantly scaled down (ClassAct; Fig. \ref{fig:fig_3_generalisation}, blue curves). For example, while ViT-B scoring models yield a 31\% learner speedup, the 50\x \ smaller ViT-Ti scoring models still provide a 26\% speedup. We pushed this logic by using even smaller scoring models (the ViT-Mu family which we introduce, see Appendix
) 
which are up to 1000\x \ smaller than the learner. Despite this, prioritizing data based on their scores yields non-negligible speedups (\eg 16\% for the smallest actors we consider; Figure \ref{fig:fig_3_generalisation}, top right). 

These experiments demonstrate that, with the appropriate scoring criterion, online and reference models can be significantly downscaled and still produce comparable gains to larger models, with learner efficiency degrading gracefully with the actor overhead (\ie the cost of the reference model and data scoring). As a result, our method ClassAct quickly becomes FLOP positive as the online + reference models are downscaled (Figure \ref{fig:fig_3_generalisation}, bottom right), while at the same time producing speed-ups in wall-clock time for a given learner batch size. 

Together, our results expose a pareto front across which to determine an optimal context-specific data selection strategy (Figure \ref{fig:bar_plot_amortization}). Where pre-trained models are available, some of the cost of larger data selection policies can be discounted. If savings in wall-clock time supersede the associated cost of scoring, large models can be tolerated for data selection. Reference model costs can also be amortized across many training runs by appending scores to `foundation datasets'. However, in the case where no component of the framework can amortized (as in the case of large-scale pretraining), prioritizing data with small ClassAct models can deliver large savings in total computation.

\subsection{Generalising neural scaling laws to the active-learning setting}

We next investigated the scaling behaviour of ClassAct by experimenting over large learner compute budgets (Fig. \ref{fig:fig_1_scaling}), using both ViT-Ti and ViT-S models as actors for training a ViT-L. Predictably, the ViT-S produced larger, although marginal gains over the ViT-Ti actors when not accounting for scoring FLOPs (Figure \ref{fig:fig_1_scaling}, left). However, when accounting for total FLOPs, the difference was less pronounced (Figure \ref{fig:fig_1_scaling}, right). Our results generalize large scale uniform sampling scaling laws such as uncovered by \cite{kaplan2020scaling} for LLMs and reproduced for large vision transformers by \cite{zhai2022scaling} to the case of non-uniform sampling. For the first time, we demonstrate that these scaling laws can be shifted in our favour by selecting data using general model-based scoring heuristics.

\subsection{Training the reference model in parallel} \label{sec:one_pass}

The reference model needs to be trained if none is already available. This two-step process adds complexity to active model training, especially if using large scale infrastructure. However, an interesting consequence of down-scaling the reference model is that both inference passes and gradients can be computed over a much larger batch than can be computed on the learner. 
In theory, this would mean that the small reference model could instead be trained \textit{online}, in parallel with the large learner and small online model.

We confirmed our hypothesis by running an experiment in which we trained our reference model on a super-batch of size $B=10b$ and trained the online and learner model in sequence with the sub-batch of size $b$. To make sure the reference model quickly converges, we additionally set the learning rate to double that of the online and learner models (this would cause instability for the learner and online models because of the additional variance from the smaller batch). We also verified that training the reference model on a held-out set of data performed equally in our experiments to reference models trained on the same data as the learner model \cite{mindermann2022prioritized}. Our `one-pass' setup, \textit{Online-ClassAct}, produces the same performance as the pre-trained ClassAct pipeline in our Ti-trains-B experiments (Table \ref{tab:one_pass_rho}). 
Pseudocode is shown in appendix Algorithm 1.

We have shown that by \textit{decoupling the scoring models from the learner model entirely}, it is possible to significantly downscale the scoring models with minor degradation to performance (see Table \ref{tab:one_pass_rho}). Unlike RHO, which can train a large `learner' model with a small `reference' model, we introduce a third `online' model, with the same architecture and parameter count as the reference model, enabling the reduction of actor computation (see 
Table \ref{tab:one_pass_rho}).

\input{tables/table2}

\begin{figure}
    \centering
    \includegraphics[width=1.0\linewidth]{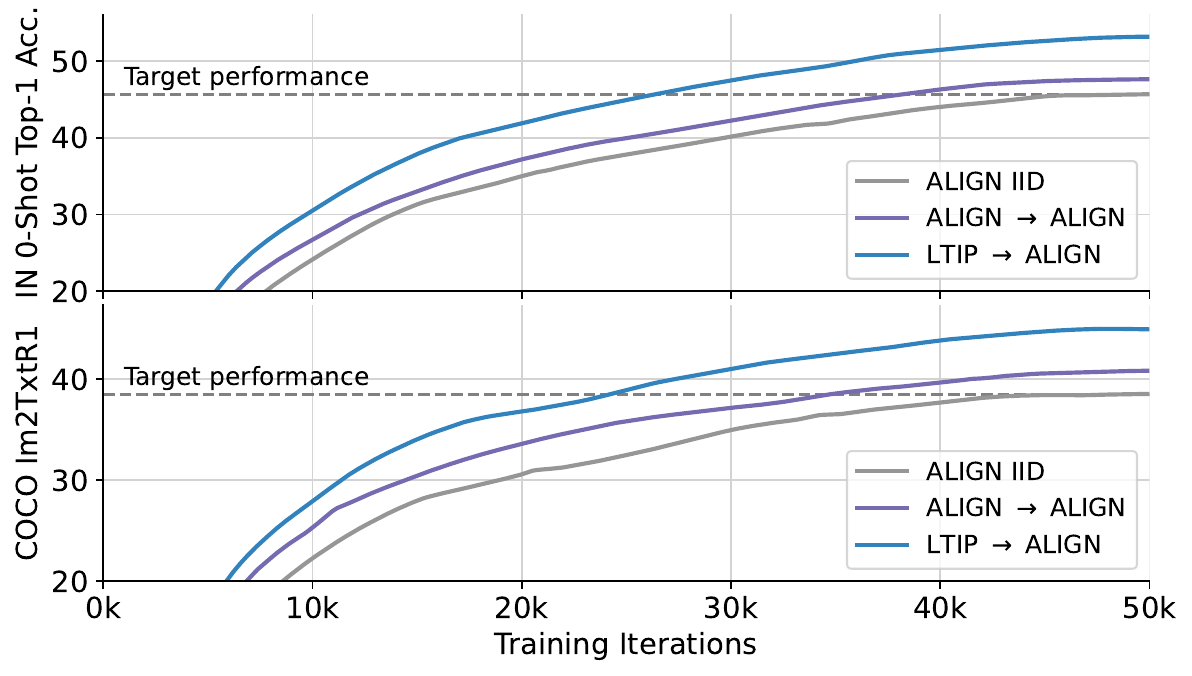}
    \caption{\textbf{Reference policies generalize across tasks}. We train a ViT-B reference model on either ALIGN or LTIP, then use it to train a second ViT-B learner on ALIGN with ActiveCLIP.  
    The biggest gains
    were found with an LTIP reference model, despite it needing to perform out-of-domain generalization. In 50k iterations, ActiveCLIP selects the 800M ``cleanest'' examples from the ALIGN dataset, whose size is 1.6B in total.  
    \vspace{-1em}}
    \label{fig:fig_4_crossmodal}
\end{figure}

\subsection{ActiveCLIP: active multimodal learning}

We have so far demonstrated that large scale image classifiers can be trained with lower total compute by actively selecting the data used for training. However, classification has largely been superseded as a large scale pre-training method by CLIP-style multimodal training \cite{radford2021learning}. Figure \ref{fig:fig_4_crossmodal} demonstrates that our CLIP-adapted active learning method \textit{ActiveCLIP} (see Methods) produces similar speedups in terms of learner computation as observed in JFT classification. Specifically, we find that prioritized sampling with learnability scores accelerates multimodal pre-training by 18-48\%, depending on the evaluation metric (ImageNet zero-shot accuracy or COCO retrieval) and reference model configuration, which we explore below. 

\begin{figure}
    \centering
    \includegraphics[width=1.0\linewidth]{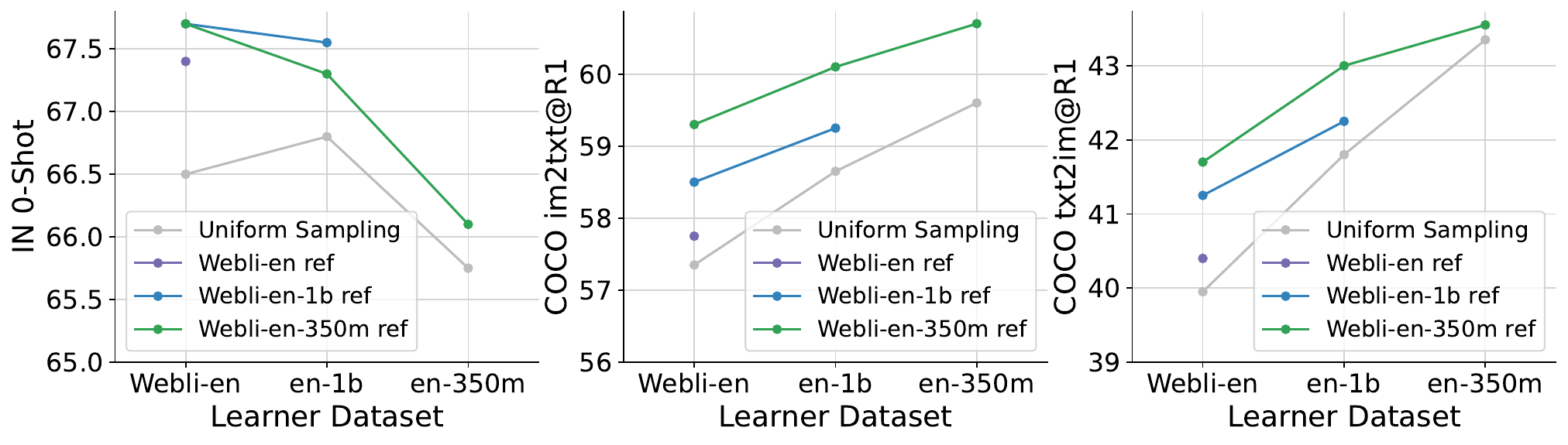}
    \caption{\textbf{Reference models trained on curated datasets are powerful data selectors}. Active data selection using \textit{ActiveSigLIP} with reference models trained on increasingly curated subsets of Webli. Webli-350M reference models are consistently more effective than those trained on raw or 1B subsets. \vspace{-1em}}
    \label{fig:fig_n_webli}
\end{figure}

\subsection{Policy generalization across tasks}

To fully leverage off-the-shelf pre-trained image models for training (Figure \ref{fig:bar_plot_amortization}), our results up to now suggest that the reference model should be trained on the same task that the learner model is being trained for. In Figure \ref{fig:fig_4_crossmodal} (see Table \ref{tab:cross_modal_contrastive} for full results), we show that we can in fact pre-train our reference models on related but distinct datasets. Moreover results suggest that there may even be a benefit to cross-training in some-cases; an ALIGN$\rightarrow$ALIGN reference$\rightarrow$learner model combination (`in-domain' ActiveCLIP) produced similar speedups to ClassAct on JFT. However, these gains were greatly surpassed by an LTIP$\rightarrow$ALIGN combination. One possibility is that LTIP is less noisily labeled such that the scoring policies it produces are `cleaner' - i.e. more able to filter for `clean' data. The corollary may also be true; active data selection appears to greatly improve the utility of ALIGN, suggesting that it contains a large proportion of data that is of low quality for training.

We observed the same effect when training with ActiveSigLIP on the large scale Webli dataset \cite{chen2022pali} (results summarised in Table \ref{tab:mulimodal_results}). We trained reference models on both the raw 4B set as well as two extensively curated 1B / 350M subsets, which were then used to train subsequent learner models (Fig. \ref{fig:fig_n_webli}). In all-but-one cases, filtering data with the 350M-trained referenced model produced the best results when transferring to 350M, 1B and 4B learner datasets. Notably, the significant gains observed over training with uniform sampling on the 350M subset suggest that our method can still improve over methods that pre-filter datasets once but then train with uniform samples \cite{gadre2023datacomp}.

\subsection{Comparison to prior multimodal art}

We leverage the insight that pre-training a reference model on clean data can facilitate learning on larger, noisy data for training our final model. Here, we train a reference model on LTIP, then use it to train a new model on the much larger mixture of LTIP and ALIGN, following \cite{alayrac2022flamingo}, for a total of 3 and 8 billion training examples. Table \ref{tab:mulimodal_results} shows that in this training regime, \textit{ActiveCLIP} surpasses models trained with significantly more data on ImageNet 0-Shot classification and COCO retrieval. Finally, we find that our active learning method is complementary to recent advances in multimodal learning: ActiveSigLIP significantly improves uniformly-trained SigLIP \cite{zhai2023sigmoid} in both COCO retrieval metrics.

\begin{table}[]
\centering
\small
\setlength\tabcolsep{3.5pt} 
\begin{tabular}{lccccc}
  &  & IN-1K & \multicolumn{2}{c}{COCO} \\ 
\cmidrule(lr){3-3} \cmidrule(lr){4-5}
 Method  & Train ex. & ZS Top-1 & im2txt & txt2im \\
 \hline 

CLIP & 13B         & 68.3          & 52.4                 & 33.1                 \\
EVA-CLIP   & \hspace{2em}3B\textcolor{gray}{$+$2B}          & \textcolor{gray}{\hspace{0.4em}69.7$^\dagger$}          & \multicolumn{1}{l}{} & \multicolumn{1}{l}{} \\
\rowcolor{ourbg} ActiveCLIP & 3B & \textbf{71.3} &  \textbf{57.7}        & \textbf{43.0}     \\ 
\vspace{-0.5em} \\ 
OpenCLIP & 34B         &     70.2      & 59.4                 & 42.3                 \\

EVA-CLIP  & \hspace{2em}8B\textcolor{gray}{$+$2B}          & \textcolor{gray}{\hspace{0.4em}74.7$^\dagger$}          & 58.7 & 42.2 \\

\rowcolor{ourbg} ActiveCLIP & 8B & \textbf{72.2} &  \textbf{60.7}        & \textbf{44.9} \\

\vspace{-0.5em} \\

 SigLIP & 3B & 72.1 & 60.7 & 42.7 \\
\rowcolor{ourbg} ActiveSigLIP & 3B & 72.0  & \textbf{63.5} & \textbf{45.3} \\

\end{tabular}
\vspace{1em}
\caption{
\textbf{Comparison of ActiveCLIP to public multimodal pre-taining methods}. ActiveCLIP outperforms models trained with the same or more data (CLIP, \cite{radford2021learning}; EVA-CLIP, \cite{sun2023eva}; and OpenCLIP, \cite{ilharco_gabriel_2021_5143773}). ActiveSigLIP produced significant gains over the baseline SOTA performance of SigLIP \cite{zhai2023sigmoid}. $^\dagger$benefits from additional ImageNet21K pretraining (+2B training examples). All ActiveCLIP/SigLIP models use a reference model trained on LTIP to guide learning on a mixture of ALIGN and LTIP.  \vspace{-2em}
}

\label{tab:mulimodal_results}
\end{table}

%% file: tables/table1.tex
\begin{table*}[t]
\small
\centering
\setlength{\tabcolsep}{8pt}

\begin{tabular}{c c c c c | c c} 
& \multicolumn{3}{c}{ViT model capacity} & \multicolumn{3}{c}{\hspace{7em} Speed-up} \\ 
\cmidrule(lr){2-4} \cmidrule(lr){6-7}
 Method & Reference & Online & Learner & Reference Type & \makecell{Learner\\speedup} & \makecell{Compute\\speedup} \\ [0.5ex] 
 \hline 
 Uniform Sampling (ViT-B) & & & B & & 0\% & 0\% \\ 
 RHO & Tiny & B & B & Held-out, fixed & 0\% & - 79\% \\ 
 ClassAct-HO                                  & Tiny & Tiny & B & Held-out, fixed & 18\% & 3\% \\ 
\rowcolor{ourbg} ClassAct & Tiny & Tiny & B & In-domain, fixed & 18\% &  3\% \\ 
 ClassAct-Online & Tiny & Tiny & B & Trained online & 17\% & 2\% \\ 
\end{tabular}
\vspace{1em}
\caption{\textbf{Simplifying and accelerating the computation of learnability scores.} Relative to RHO \cite{mindermann2022prioritized}, \ourmethod \ makes two changes: replacing the reference model with one trained in-domain (removing the need for bespoke held-out sets), and dramatically reducing capacity of the online actor models used for scoring examples. All experiments were conducted on 3 epochs of one-half of JFT to enable the held-out ablations. RHO with a small reference model did not produce a learner speedup in our experiments. \vspace{-1em}}
\label{tab:one_pass_rho}
\end{table*}

%% file: tables/table2.tex
\begin{table*}[t]
\centering
\small
\begin{tabular}{cccccccccc}
& \multicolumn{2}{c}{ImageNet 0-Shot} &
&
\multicolumn{2}{c}{COCO (im2txtR1)} &
&
\multicolumn{2}{c}{COCO (txt2imR1)}
&
\multicolumn{1}{l}{\textit{(Learner)}} 
\\ \cline{2-3} \cline{5-6} \cline{8-9}
\multicolumn{1}{c|}{\multirow{2}{*}{\makecell{Speed-up \%\\(vs. Uniform)}}} &
\multicolumn{1}{c|}{\spdupgrad{23.0}} &
\multicolumn{1}{c|}{\spdupgrad{48.1}} &
\multicolumn{1}{c|}{} &
\multicolumn{1}{c|}{\spdupgrad{28.6}} &
\multicolumn{1}{c|}{\spdupgrad{48.0}} &
\multicolumn{1}{c|}{} &
\multicolumn{1}{c|}{\spdupgrad{26.8}} &
\multicolumn{1}{c|}{\spdupgrad{27.0}} &
ALIGN
\\ \cline{2-3} \cline{5-6} \cline{8-9}
\multicolumn{1}{c|}{} &
\multicolumn{1}{c|}{\spdupgrad{0.0}} &
\multicolumn{1}{c|}{\spdupgrad{16.3}} &
\multicolumn{1}{c|}{} &
\multicolumn{1}{c|}{\spdupgrad{18.8}} &
\multicolumn{1}{c|}{\spdupgrad{22.1}} &
\multicolumn{1}{c|}{} & \multicolumn{1}{c|}{\spdupgrad{28.2}} &
\multicolumn{1}{c|}{\spdupgrad{17.5}} &
LTIP
\\ \cline{2-3} \cline{5-6} \cline{8-9}
& & & & & & & &                                 
& \multicolumn{1}{l}{} \\ \cline{2-3} \cline{5-6} \cline{8-9}
\multicolumn{1}{c|}{\multirow{2}{*}{\makecell{Performance\\(vs. Uniform)}}} &
\multicolumn{1}{c|}{47.8 (\perfgrad{2.2})} &
\multicolumn{1}{c|}{53.2 (\perfgrad{7.4})} &
\multicolumn{1}{c|}{} &
\multicolumn{1}{c|}{40.9 (\perfgrad{2.4})} &
\multicolumn{1}{c|}{44.8 (\perfgrad{6.3})} &
\multicolumn{1}{c|}{} & \multicolumn{1}{c|}{27.3 (\perfgrad{1.8})} &
\multicolumn{1}{c|}{30.5 (\perfgrad{5.0})} &
ALIGN
\\ \cline{2-3} \cline{5-6} \cline{8-9}
\multicolumn{1}{c|}{} &
\multicolumn{1}{c|}{46.5 (\perfgrad{-0.1})} &
\multicolumn{1}{c|}{47.2 (\perfgrad{0.6})} &
\multicolumn{1}{c|}{} &
\multicolumn{1}{c|}{45.6 (\perfgrad{1.2})} &
\multicolumn{1}{c|}{45.4 (\perfgrad{1.0})} &
\multicolumn{1}{c|}{} & \multicolumn{1}{c|}{31.8 (\perfgrad{1.7})} &
\multicolumn{1}{c|}{31.0 (\perfgrad{0.9})} & 
LTIP
\\ 
\cline{2-3} \cline{5-6} \cline{8-9}
\vspace{-4pt}
\\
\multicolumn{1}{l}{\textit{(Reference)}}
&
ALIGN &
LTIP &
\multicolumn{1}{l}{}  &
ALIGN &
LTIP &
\multicolumn{1}{l}{} &
ALIGN & 
LTIP &
\multicolumn{1}{l}{}
\end{tabular}
\vspace{1em}
\caption{\textbf{Generalizing data-selection policies across datasets and tasks.} We pretrain reference models on the large but noisy ALIGN dataset, or the smaller and more curated LTIP dataset \cite{alayrac2022flamingo}. Consistent with \cite{alayrac2022flamingo}, we find training with uniform sampling on LTIP to yield stronger transfer learning results than pre-training on ALIGN. These reference models can be used very effectively for data-selection on both LTIP and ALIGN, whereas ALIGN-pretrained reference models yield more modest speedups. All models are provided with 800M training images at resolution 128\x128, speedups are shown relative to the time at which the uniform sampling baseline was reached for that evaluation metric. Colour indicates performance relative to uniform (brackets).}
\label{tab:cross_modal_contrastive}
\end{table*}

%% file: sec/5_discussion.tex
\section{Discussion}

In this work, we have presented a new method for active data selection that builds upon and simplifies the concept of `learnability'. Our experiments demonstrate that this approach can significantly reduce the computation required for large-scale pretraining, compared to training with uniform samples. To our knowledge, this is the first active learning method that is more efficient than training with uniform samples when accounting for total FLOPs, and that does not rely on hand-designed features, allowing broad application across training setups. We have validated this by showing results on classification and contrastive pre-training, and found that our data selection policies continue to produce efficiency gains in the large-scale regime and can generalize effectively across task modalities. Collectively, our experiments also illustrate a Pareto frontier that allows trading off actor/data-selection computation against savings in training iterations, suggesting an alternative path to improved performance beyond scaling training batch sizes.

This work focused on supervised pre-training for images, but further work could involve extending our method to other modalities and training schemes such as language, video, and generative modeling. An important note is that all our experiments present results from filtering only 50\% of the data; further gains may be possible by filtering more aggressively, at the cost of further overheads. In particular, aggressive data-selection coupled with efficient scoring schemes such as the ones proposed here could test the hypothesis that large-scale pretraining can benefit from exponential, rather than power-law, scaling behavior. 

\vspace{3em}

%% file: sec/6_appendix.tex
\newpage

\appendix

\section{Supplementary}

\subsection{Details of total compute calculations}
\label{app:flops}

The average FLOPs per learner update for an IID training run is given by:

$$
C_{\text{IID}} = 3 F_{\text{learn}}
$$
where $F_{\text{learn}}$ is the cost of a single learner inference pass and a gradient update costs $\sim$3$\times$ inference passes \cite{jouppi2017datacenter}. The average FLOPs per learner update for easy-reference prioritisation is:

$$
C_{\text{easy\_ref}} =
\big(
3 F_{\text{learn}} +
\frac{F_{\text{ref}}}{\text{SPI}}
\big) \beta + 
3 F_{\text{ref}}
$$
where $\text{SPI} = b/B$ ($=0.5$ for our experiments) and $\beta$ is the learner speedup vs. uniform sampling. The average FLOPs per learner update for RHO is:

$$
C_{\text{RHO}} =
\big(
3 F_{\text{learn}} +
(F_{\text{learn}} + \frac{F_{\text{ref}})}{\text{SPI}} + 
\big) \beta + 
3 F_{\text{ref}}
$$
And the average FLOPs per update for \textit{ActiveCLIP} / \textit{ClassAct} is:

$$
C_{\text{ClassAct / ActiveCLIP}} =
\big(
3 F_{\text{learn}} +
2\frac{F_{\text{ref}}}{\text{SPI}}
\big) \beta + 
3 F_{\text{ref}}
$$
The conditions for compute positivity are therefore:

$$
\Big( 3 F_{\text{learn}} + \frac{F_{\text{act}}}{\text{SPI}} \Big) \beta + 3 F_{\text{ref}}
< 3 F_{\text{learn}}
$$
where $F_{\text{act}}$ is one of $F_{\text{ref}}$, $\big(F_{\text{learn}} + F_{\text{ref}} \big)$ or $2F_{\text{ref}}$ for "easy reference", RHO or ClassAct / ActiveCLIP, respectively.

\subsection{Implementation details: ClassAct}
\label{app:class}

\vspace{0.5em} \noindent \textbf{Dataset:} We pretrain on the JFT-300M dataset which contains labels for 18,292 classes collected in a semi-automated fashion. JFT labels are therefore noisy (consistently with other web-scale datasets), with approximately 20\% of the examples being mislabeled. In the ``standard'' setting both reference models and subsequent ClassAct models are trained on the full JFT-300M dataset. 

\vspace{0.5em} \noindent \textbf{Architectures:} We use standard vision transformers trained with 224\x224 image resolution and a patch size of 16\x16. We consider the standard ViT-Base and ViT-Large variants as learners in our in our main study, and smaller-scale variants (ViT-S and ViT-Tiny) when reducing the scoring-model computation. We introduce even smaller variants to scale down actor computation further, and detail the configurations of this ``ViT-Micro'' family in section \ref{sec:app_small_vits}.

\vspace{0.5em} \noindent \textbf{Learner Training:} We train visual classifiers with softmax cross-entropy and label smoothing of 0.1. We use the AdamW optimizer \cite{loshchilov2019decoupled} with a cosine learning-rate decay and warmup period of 10k iterations. The maximum learning rate is 0.001 and the weight decay is $10^{-3}$.

\vspace{0.5em} \noindent \textbf{Evaluation:} We evaluate the performance of the learned model by applying it to a held-out set of JFT images, and measuring top-1 accuracy. Previous work has found that in-domain performance on JFT is highly predictive of out-of-distribution generalization \cite{zhai2022scaling}, hence we leave the study of transfer performance to the multimodal setting (Sec. \ref{app:clip}).

\subsection{Implementation details: ActiveCLIP}
\label{app:clip}

\vspace{0.5em} \noindent \textbf{Datasets:}
We use paired image-text datasets for contrastive pretraining. We experiment with ALIGN, composed of 1.8B image-text pairs, LTIP, introduced in \cite{alayrac2022flamingo} and consisting of 312M higher-quality images with longer descriptions, and JFT-300M which despite being a classification dataset can be used for contrastive learning by using the labels as the paired text for each image \cite{yu2022coca}.

\vspace{0.5em} \noindent \textbf{Architectures:} The model consists of a vision and text encoder in the standard contrastive learning setup introduced in \cite{radford2021learning}. We a vision transformer as the visual encoder and a text transformer encoder. We use the Base variant for both the vision and text transformers on the learner, following \cite{dosovitskiy2020image}. The smaller variants used by scoring models are the same as in the classification setup described above. 

\vspace{0.5em} \noindent \textbf{Learner Training:} As described in Section \ref{sec:meth-losses} we optimize the model using the standard contrastive loss. We train the entire model with the AdamW optimiser and decay the learning rate using a cosine schedule (following a linear warmup period) with a maximum value of $10^{-3}$. We also use a learnable temperature parameter as described in \cite{radford2021learning} and clip the gradients to a maximum global norm of 1.0. We use a batch size of 16,384 on the learner for each dataset. 

\vspace{0.5em} \noindent \textbf{Evaluation:}
We evaluate the model’s zero-shot transfer results on standard multimodal benchmarks---image-text retrieval on COCO \cite{lin2014microsoft} and zero-shot image classification on ImageNet. For COCO, we use the standard technique of comparing image and text representations across the dataset to find the most similar text (or image) for each image (or text). For ImageNet, we precompute the representations for all predefined labels and calculate the most similar label for each image. For zero-shot classification, we also apply the prompt engineering described in \cite{radford2021learning} using the released prompts\footnote{\url{https://github.com/openai/CLIP/blob/main/notebooks/Prompt\_Engineering\_for\_ImageNet.ipynb}}. We average the embeddings for each label across all prompts before calculating similarity with the images. 




\subsection{Online ClassAct / ActiveCLIP}

ClassAct / ActiveCLIP can also be run in a single training loop, by training the reference model on the super-batch:

\begin{algorithm}
\small
\caption{Online ClassAct / ClassCLIP}
\begin{algorithmic}[1]
\State \textbf{Input:} Randomly initialized learner model $\theta_l$, small online and reference models $\theta_o$ and $\theta_r$. Models use loss $\ell_\textrm{act}$ for scoring data and $\ell_\textrm{learn}$ for computing updates. Dataset $\mathcal{D}$, batch size $B$, sub-batch size $b<B$. Fast and slow learning rates $\alpha$ and $\beta$.
\While{training}
    \State $X \sim \mathcal{D}$, where $|X| = B$ \Comment{Sample uniformly}
    \State $S = \ell_{\text{act}}(X; \theta_o) - \ell_{\text{act}}(X; \theta_r)$ \Comment{Get scores}
    \State $\theta_r \leftarrow \text{Adam}_{\alpha}[\nabla_{\theta_r} \ell_{\text{learn}}(X|\theta_l)]$ \Comment{Update ref. model}
    \State $I \sim \text{SoftMax}(S)$, where $|I| = b$ \Comment{Sample indices}
    \State $Y = X[I]$ \Comment{Collect sub-batch}
    \State $\theta_l \leftarrow \text{Adam}_{\beta}[\nabla_{\theta_l} \ell_{\text{learn}}(Y|\theta_l)]$ \Comment{Update learner model}
    \State $\theta_o \leftarrow \text{Adam}_{\beta}[\nabla_{\theta_o} \ell_{\text{learn}}(Y|\theta_o)]$ \Comment{Update online model}
\EndWhile
\end{algorithmic}
\label{alg:pretrained}
\end{algorithm}

We found that $B=10b$ (SPI=0.1) was sufficient to reproduce the pre-trained reference (Algorithm \ref{alg:classact_algo}) case where $B=2b$ (SPI=0.5), as described Section \ref{sec:one_pass}. It is possible that the super-batch size could be shrunk, but we did not evaluate SPI values in between these values.

\subsection{Analyzing learnability scores}

Here we examine learnability scores in the case of small differences between the online and reference models $\theta^t, \theta^*$. In this case, we can apply a Taylor expansion of the reference-model loss: 
\begin{equation}
    \ell(\vx_i | \theta^*) \approx  \ell(\vx_i | \theta^t) + (\theta^* - \theta^t) \cdot \nabla_{\theta} \ell( \vx_i |\theta^t)
\end{equation}
and learnability scores simplify to the alignment between the gradient of the loss with respect to the online parameters, and the difference between online and reference parameters:
\begin{align}
\label{eq:learn_analysis}
s^\textrm{learn}(\vx_i | \theta^t, \theta^*)
& =  \ell(\vx_i | \theta^t) - \ell(\vx_i | \theta^*) \\ 
& \approx  (\theta^t - \theta^*) \cdot \nabla_{\theta} \ell( \vx_i |\theta^t).
\end{align}
This yields a simple explanation for the relevance of learnability scores: data points whose gradient aligns well with the ``direction of travel'' (\ie the difference between the current model state and the fully-trained one) will be prioritized. We can further simplify this expression in the case where the reference model is the result of a single, global update with respect to gradients from a batch of data $\{\vx_j\}$:
\begin{align}
    \theta^* & = \theta^t - \lambda \nabla_{\theta} \ell( \{ \vx_j \} |\theta) \\ 
    s^\textrm{learn}(\vx_i | \theta^t, \theta^*) & \approx \lambda \nabla_{\theta} \ell( \vx_i |\theta^t) \cdot \nabla_{\theta} \ell( \{\vx_j\} | \theta^t)
\end{align}
In this case, the learnability of an example reduces to the alignment of its gradient to those of the batch. In particular, these scores will de-prioritize examples which are trivially solved (small gradients) or noisy (mis-aligned with the batch gradient). 

Finally, we note that learnability scores reduce to gradient-norm prioritization \cite{paul2021deep} if we make the further (more stringent) assumption that the reference model is the result of a single update with respect to \textit{this example's gradient only}:
\begin{align}
    \theta^* & = \theta^t - \lambda \nabla_{\theta} \ell( \vx_i |\theta) \\ 
    s^\textrm{learn}(\vx_i | \theta^t, \theta^*) & \approx s^\textrm{grad}(\vx_i | \theta^t) =  \lambda \left|\left| \nabla_{\theta} \ell( \vx_i |\theta^t) \right|\right|^2
\end{align}
While this prioritization effectively discards examples with small gradients, it does not benefit from the denoising properties of the more general formulation of learnability, as noisy examples can exhibit large gradients. 

\subsection{Learning infrastructure} \label{sec:distributed_setup}

\begin{figure}[t]
  \centering
  \includegraphics[scale=0.38]{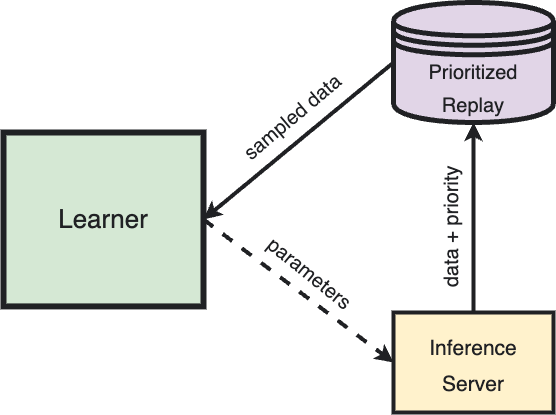}
  \caption{The main nodes of our distributed learning infrastructure are: (1) a learner that continuously updates model parameters based on sampled data, (2) an inference server which computes priorities of uniformly sampled data, and (3) a prioritized replay which receives data and priorities from the inference server and samples prioritized data for learning. The critical aspect of our system is adaptive data prioritization which is enabled by keeping inference parameters in sync with the learner. We employ separate prioritized replay and inference server nodes per dataset. We also employ run loop nodes (not shown in the diagram) for reading and writing data. Run loop nodes can run as threads on the inference server node or remotely, to prevent extra load on inference servers.}
  \label{fig:learning_infra}
\end{figure}

Our learning infrastructure Figure \ref{fig:learning_infra} draws inspiration from distributed reinforcement learning (DRL) \cite{espeholt2018impala}. In contrast to DRL, where run loops generate data through interactions with an environment, run loops in our system are modified to read offline data. In addition to obtaining data, run loops run inference on the data via remote inference servers and write the inference outputs to prioritized replay data stores \cite{cassirer2021reverb}. The distributed nodes of our learning setup are configured and connected using \cite{yang2021launchpad}.

Data stored is sampled from the prioritized replay based on priorities determined during inference. Remote inference servers continuously run inference at specific batch sizes, in parallel to learning. Parameters of the inference servers are updated after each learning step.

This setup allows us to run ActiveCLIP at the same speed as uniform sampling experiments, even without approximating the scoring models. This distributed setup may be advantageous over the sequential online batch selection setup (where the actors and learners run on the same hardware) in cases where time, not compute, is the bottleneck; adding computation to parallelize the data filtration is an alternative to increasing the batch size, which is known to saturate \cite{zhai2023sigmoid}.

For stability, we control the ratio of data sampled to data inserted in all of our experiments. For example, a samples-per-insert (SPI) ratio of 0.5 means that half of the inserted data does not get sampled, ensuring the learner focuses on the most relevant data. Each data item is sampled at most once. These two constraints impose minimum inference requirements to fully saturate the learner, which we separately tune for each experiment by scaling up the topology of the inference server.

Our infrastructure is designed to work with multiple datasets. In this case, we use separate remote inference server, run loops, and prioritized replay for each dataset, and set the batch size per dataset.

To continuously assess the performance of our models, we add data evaluator nodes which measure performance on held-out datasets while periodically pulling the latest parameters from the learner node.

\subsection{Small ViT models} \label{sec:app_small_vits}

We extended the suite of ViT models presented in \cite{zhai2022scaling} by further scaling down the ViT-Ti model as shown in Table \ref{tab:small_vits}.

\begin{table}[t]
\footnotesize
\centering
\begin{tabular}{ccccccc}
Name     & Width & Depth & MLP  & Heads & Mio- & GFLOPs \\
     &  &  &   &  & -Params & $224^2$ \\
\hline
L        & 1024  & 24    & 4096 & 16    & 304        & 61.6           \\
B        & 768   & 12    & 3072 & 12    & 86         & 17.6           \\
S        & 384   & 12    & 3072 & 12    & 22         & 4.6            \\
Ti       & 192   & 12    & 768  & 3     & 5.6        & 1.3            \\
Mu\_6-3  & 192   & 6     & 768  & 3     & 2.8        & 1.24           \\
Mu\_12-2 & 128   & 12    & 512  & 2     & 2.5        & 1.23           \\
Mu\_6-2  & 128   & 6     & 512  & 2     & 0.66       & 0.48           \\
Mu\_12-1 & 64    & 12    & 256  & 1     & 1.3        & 0.23           \\
Mu\_6-1  & 64    & 6     & 256  & 1     & 0.36       & 0.203           \\
Mu\_3-1  & 64    & 3     & 256  & 1     & 0.22       & 0.065           \\
Mu\_2-1  & 64    & 2     & 256  & 1     & 0.16       & 0.033           \\
Mu\_1-1  & 64    & 1     & 256  & 1     & 0.12       & 0.011  
\end{tabular}
\caption{\textbf{ViT model details}. Configurations, parameter counts, and FLOPs associated with small ViT models.}
\label{tab:small_vits}
\end{table}

\section{Appendix: supplementary results}

\subsection{FLOP efficient contrastive training}

Our previous results (Table \ref{tab:cross_modal_contrastive}) showed that it is possible to improve the performance of contrastve pre-training using active data selection via ActiveCLIP. However, these results used a ViT-B policy (reference and online models) to train a ViT-B learner. To explore whether similar gains could be achieved with fewer FLOPs, we repeated out analyses using ViT-Ti policies (Table \ref{tab:mulimodal_results_ti}). As in our ClassAct JFT experiments, our results show that performance degrades marginally, eve though the ViT-Ti policy requires 50x fewer FLOPs.

\begin{table}[]
\centering
\small
\begin{tabular}{lccccc}
  &  & IN-1K & \multicolumn{2}{c}{COCO} \\ 
\cmidrule(lr){3-3} \cmidrule(lr){4-5}
 Method  & Train ex. & ZS Top-1 & im2txt & txt2im \\
 \hline 

CLIP \cite{radford2021learning} & 13B         & 68.3          & 52.4                 & 33.1                 \\
EVA-CLIP \cite{sun2023eva}    & 3B          & 69.7          & \multicolumn{1}{l}{} & \multicolumn{1}{l}{} \\
SigLIP \cite{zhai2023sigmoid}      & 3B          & 68.8 & 56.7        & 40.2                 \\
\rowcolor{ourbg} ActiveCLIP-B & 3B & \textbf{70.3} &  \textbf{57.8}        & \textbf{42.6} \\
\rowcolor{ourbg} ActiveCLIP-Ti & 3B & \textbf{69.9} &  \textbf{57.2}        & \textbf{42.4}     
\end{tabular}
\caption{\textbf{Comparison of \textit{ActiveCLIP} to public multimodal pre-taining methods} Extension of Table \ref{tab:cross_modal_contrastive}, replacing the ViT-B reference and online models with a ViT-Ti. The ViT-Ti policy marginally underperforms the ViT-B policy, but is also SOTA vs. previous baselines.}
\label{tab:mulimodal_results_ti}
\end{table}

We also repeated this ablation for our out-of-domain contrastive pre-training experiments shown in Fig. \ref{fig:fig_4_crossmodal}). Again, ViT-Ti policies produces similar speedups and were also compute-positive, mirroring our ClassAct JFT experiments (Figure \ref{fig:fig_1_scaling}), \ref{fig:fig_3_generalisation}).

\begin{table*}[t]
\small

\resizebox{\textwidth}{!}{
\begin{tabular}{c c|ccc|ccc|ccc}
 & &
 \multicolumn{3}{c|}{\underline{ImageNet 0-Shot} } &
 \multicolumn{3}{c|}{\underline{COCO (im2txt)} } &
 \multicolumn{3}{c}{\underline{COCO (txt2im)} } \\ 
\multirow{2}{*}{Method} & Ref. &
Speed & FLOP & Top 1&
Speed & FLOP & \multirow{2}{*}{R@1} &
Speed & FLOP & \multirow{2}{*}{R@1} \\
& model &
-up & saving & Acc. &
-up & saving &  &
-up & saving &  \\
\hline


\hline
IID &  & 0.0\% & 0.0\% & 0.386 & 0.0\% & 0.0\% & 0.256 & 0.0\% & 0.0\% & 0.457 \\
ActiveCLIP & ViT-Ti & 43.3\% & 26.2\% & 0.428 & 38.7\% & 20.8\% & 0.292 & 38.3\% & 20.3\% & 0.509 \\
ActiveCLIP & ViT-B & 51.0\% & -163.4\% & 0.449 & 48.3\% & -172.4\% & 0.305 & 47.3\% & -175.6\% & 0.532 \\

\hline
\vspace{0.2em}
\end{tabular}
}

\caption{\textbf{Compute-positive contrastive pre-training}. Reproduction of experiments from Figure \ref{fig:fig_4_crossmodal}, replacing the ViT-B reference and online models with ViT-Ti. As in our ClassAct JFT experiments (Figure \ref{fig:fig_1_scaling}), \ref{fig:fig_3_generalisation}), the ViT-Ti policy produced nearly the same learner speedups as the ViT-B policy, but unlike the ViT-B policy also produced net FLOP savings. In all cases the reference model is trained on LTIP and the learner is trained on ALIGN.}
\end{table*}




%% file: main.bbl
\begin{thebibliography}{10}
\providecommand{\url}[1]{\texttt{#1}}
\providecommand{\urlprefix}{URL }
\providecommand{\doi}[1]{https://doi.org/#1}

\bibitem{abbas2023semdedup}
Abbas, A., Tirumala, K., Simig, D., Ganguli, S., Morcos, A.S.: Semdedup:
  Data-efficient learning at web-scale through semantic deduplication. arXiv
  preprint arXiv:2303.09540  (2023)

\bibitem{alayrac2022flamingo}
Alayrac, J.B., Donahue, J., Luc, P., Miech, A., Barr, I., Hasson, Y., Lenc, K.,
  Mensch, A., Millican, K., Reynolds, M., et~al.: Flamingo: A visual language
  model for few-shot learning. Advances in Neural Information Processing
  Systems  (2022)

\bibitem{brown2020language}
Brown, T., Mann, B., Ryder, N., Subbiah, M., Kaplan, J.D., Dhariwal, P.,
  Neelakantan, A., Shyam, P., Sastry, G., Askell, A., et~al.: Language models
  are few-shot learners. Advances in Neural Information Processing Systems
  (2020)

\bibitem{campbell2018bayesian}
Campbell, T., Broderick, T.: Bayesian coreset construction via greedy iterative
  geodesic ascent. In: International Conference on Machine Learning. pp.
  698--706. PMLR (2018)

\bibitem{cassirer2021reverb}
Cassirer, A., Barth-Maron, G., Brevdo, E., Ramos, S., Boyd, T., Sottiaux, T.,
  Kroiss, M.: Reverb: A framework for experience replay (2021)

\bibitem{chen2022pali}
Chen, X., Wang, X., Changpinyo, S., Piergiovanni, A., Padlewski, P., Salz, D.,
  Goodman, S., Grycner, A., Mustafa, B., Beyer, L., et~al.: Pali: A
  jointly-scaled multilingual language-image model. arXiv preprint
  arXiv:2209.06794  (2022)

\bibitem{chowdhery2023palm}
Chowdhery, A., Narang, S., Devlin, J., Bosma, M., Mishra, G., Roberts, A.,
  Barham, P., Chung, H.W., Sutton, C., Gehrmann, S., et~al.: Palm: Scaling
  language modeling with pathways. Journal of Machine Learning Research
  \textbf{24}(240),  1--113 (2023)

\bibitem{coleman2019selection}
Coleman, C., Yeh, C., Mussmann, S., Mirzasoleiman, B., Bailis, P., Liang, P.,
  Leskovec, J., Zaharia, M.: Selection via proxy: Efficient data selection for
  deep learning. arXiv preprint arXiv:1906.11829  (2019)

\bibitem{dosovitskiy2020image}
Dosovitskiy, A., Beyer, L., Kolesnikov, A., Weissenborn, D., Zhai, X.,
  Unterthiner, T., Dehghani, M., Minderer, M., Heigold, G., Gelly, S., et~al.:
  An image is worth 16x16 words: Transformers for image recognition at scale.
  In: International Conference on Learning Representations (2021)

\bibitem{espeholt2018impala}
Espeholt, L., Soyer, H., Munos, R., Simonyan, K., Mnih, V., Ward, T., Doron,
  Y., Firoiu, V., Harley, T., Dunning, I., et~al.: Impala: Scalable distributed
  deep-rl with importance weighted actor-learner architectures. In:
  International conference on machine learning. pp. 1407--1416. PMLR (2018)

\bibitem{feldman2020does}
Feldman, V.: Does learning require memorization? a short tale about a long
  tail. In: Proceedings of the 52nd Annual ACM SIGACT Symposium on Theory of
  Computing. pp. 954--959 (2020)

\bibitem{gadre2023datacomp}
Gadre, S.Y., Ilharco, G., Fang, A., Hayase, J., Smyrnis, G., Nguyen, T.,
  Marten, R., Wortsman, M., Ghosh, D., Zhang, J., et~al.: Datacomp: In search
  of the next generation of multimodal datasets. arXiv preprint
  arXiv:2304.14108  (2023)

\bibitem{gunasekar2023textbooks}
Gunasekar, S., Zhang, Y., Aneja, J., Mendes, C.C.T., Del~Giorno, A., Gopi, S.,
  Javaheripi, M., Kauffmann, P., de~Rosa, G., Saarikivi, O., et~al.: Textbooks
  are all you need. arXiv preprint arXiv:2306.11644  (2023)

\bibitem{har2005smaller}
Har-Peled, S., Kushal, A.: Smaller coresets for k-median and k-means
  clustering. In: Proceedings of the twenty-first annual symposium on
  Computational geometry. pp. 126--134 (2005)

\bibitem{hessel2021clipscore}
Hessel, J., Holtzman, A., Forbes, M., Bras, R.L., Choi, Y.: Clipscore: A
  reference-free evaluation metric for image captioning. arXiv preprint
  arXiv:2104.08718  (2021)

\bibitem{hoffmann2022training}
Hoffmann, J., Borgeaud, S., Mensch, A., Buchatskaya, E., Cai, T., Rutherford,
  E., Casas, D.d.L., Hendricks, L.A., Welbl, J., Clark, A., et~al.: Training
  compute-optimal large language models. In: Advances in Neural Information
  Processing Systems (2022)

\bibitem{ilharco_gabriel_2021_5143773}
Ilharco, G., Wortsman, M., Wightman, R., Gordon, C., Carlini, N., Taori, R.,
  Dave, A., Shankar, V., Namkoong, H., Miller, J., Hajishirzi, H., Farhadi, A.,
  Schmidt, L.: Openclip (Jul 2021). \doi{10.5281/zenodo.5143773},
  \url{https://doi.org/10.5281/zenodo.5143773}, if you use this software,
  please cite it as below.

\bibitem{jouppi2017datacenter}
Jouppi, N.P., Young, C., Patil, N., Patterson, D., Agrawal, G., Bajwa, R.,
  Bates, S., Bhatia, S., Boden, N., Borchers, A., et~al.: In-datacenter
  performance analysis of a tensor processing unit. In: Proceedings of the 44th
  annual international symposium on computer architecture. pp. 1--12 (2017)

\bibitem{kaplan2020scaling}
Kaplan, J., McCandlish, S., Henighan, T., Brown, T.B., Chess, B., Child, R.,
  Gray, S., Radford, A., Wu, J., Amodei, D.: Scaling laws for neural language
  models. arXiv preprint arXiv:2001.08361  (2020)

\bibitem{lin2014microsoft}
Lin, T.Y., Maire, M., Belongie, S., Hays, J., Perona, P., Ramanan, D.,
  Doll{\'a}r, P., Zitnick, C.L.: Microsoft coco: Common objects in context. In:
  Eur. Conf. Comput. Vis. pp. 740--755. Springer (2014)

\bibitem{lindley1956measure}
Lindley, D.V.: On a measure of the information provided by an experiment. The
  Annals of Mathematical Statistics  \textbf{27}(4),  986--1005 (1956)

\bibitem{loshchilov2015online}
Loshchilov, I., Hutter, F.: Online batch selection for faster training of
  neural networks. arXiv preprint arXiv:1511.06343  (2015)

\bibitem{loshchilov2019decoupled}
Loshchilov, I., Hutter, F.: Decoupled weight decay regularization. In:
  International Conference on Learning Representations (2019)

\bibitem{mackay1992information}
MacKay, D.J.: Information-based objective functions for active data selection.
  Neural computation  \textbf{4}(4),  590--604 (1992)

\bibitem{mahmoud2023sieve}
Mahmoud, A., Elhoushi, M., Abbas, A., Yang, Y., Ardalani, N., Leather, H.,
  Morcos, A.: Sieve: Multimodal dataset pruning using image captioning models.
  arXiv preprint arXiv:2310.02110  (2023)

\bibitem{marion2023less}
Marion, M., {\"U}st{\"u}n, A., Pozzobon, L., Wang, A., Fadaee, M., Hooker, S.:
  When less is more: Investigating data pruning for pretraining llms at scale.
  arXiv preprint arXiv:2309.04564  (2023)

\bibitem{mindermann2022prioritized}
Mindermann, S., Brauner, J.M., Razzak, M.T., Sharma, M., Kirsch, A., Xu, W.,
  H{\"o}ltgen, B., Gomez, A.N., Morisot, A., Farquhar, S., et~al.: Prioritized
  training on points that are learnable, worth learning, and not yet learnt.
  In: International Conference on Machine Learning. pp. 15630--15649. PMLR
  (2022)

\bibitem{paul2021deep}
Paul, M., Ganguli, S., Dziugaite, G.K.: Deep learning on a data diet: Finding
  important examples early in training. Advances in Neural Information
  Processing Systems  \textbf{34},  20596--20607 (2021)

\bibitem{prabhu2019sampling}
Prabhu, A., Dognin, C., Singh, M.: Sampling bias in deep active classification:
  An empirical study. arXiv preprint arXiv:1909.09389  (2019)

\bibitem{radford2021learning}
Radford, A., Kim, J.W., Hallacy, C., Ramesh, A., Goh, G., Agarwal, S., Sastry,
  G., Askell, A., Mishkin, P., Clark, J., et~al.: Learning transferable visual
  models from natural language supervision. In: International Conference on
  Machine Learning (2021)

\bibitem{schaul2015prioritized}
Schaul, T., Quan, J., Antonoglou, I., Silver, D.: Prioritized experience
  replay. arXiv preprint arXiv:1511.05952  (2015)

\bibitem{schuhmann2022laion}
Schuhmann, C., Beaumont, R., Vencu, R., Gordon, C., Wightman, R., Cherti, M.,
  Coombes, T., Katta, A., Mullis, C., Wortsman, M., et~al.: Laion-5b: An open
  large-scale dataset for training next generation image-text models. Advances
  in Neural Information Processing Systems  \textbf{35},  25278--25294 (2022)

\bibitem{schuhmann2021laion}
Schuhmann, C., Vencu, R., Beaumont, R., Kaczmarczyk, R., Mullis, C., Katta, A.,
  Coombes, T., Jitsev, J., Komatsuzaki, A.: Laion-400m: Open dataset of
  clip-filtered 400 million image-text pairs. arXiv preprint arXiv:2111.02114
  (2021)

\bibitem{settles2009active}
Settles, B.: Active learning literature survey  (2009)

\bibitem{sorscher2022beyond}
Sorscher, B., Geirhos, R., Shekhar, S., Ganguli, S., Morcos, A.: Beyond neural
  scaling laws: beating power law scaling via data pruning. Advances in Neural
  Information Processing Systems  \textbf{35},  19523--19536 (2022)

\bibitem{sun2017revisiting}
Sun, C., Shrivastava, A., Singh, S., Gupta, A.: Revisiting unreasonable
  effectiveness of data in deep learning era. In: Proceedings of the IEEE
  international conference on computer vision. pp. 843--852 (2017)

\bibitem{sun2023eva}
Sun, Q., Fang, Y., Wu, L., Wang, X., Cao, Y.: Eva-clip: Improved training
  techniques for clip at scale. arXiv preprint arXiv:2303.15389  (2023)

\bibitem{toneva2018empirical}
Toneva, M., Sordoni, A., Combes, R.T.d., Trischler, A., Bengio, Y., Gordon,
  G.J.: An empirical study of example forgetting during deep neural network
  learning. arXiv preprint arXiv:1812.05159  (2018)

\bibitem{xie2023doremi}
Xie, S.M., Pham, H., Dong, X., Du, N., Liu, H., Lu, Y., Liang, P., Le, Q.V.,
  Ma, T., Yu, A.W.: Doremi: Optimizing data mixtures speeds up language model
  pretraining (2023)

\bibitem{xie2023data}
Xie, S.M., Santurkar, S., Ma, T., Liang, P.: Data selection for language models
  via importance resampling. arXiv preprint arXiv:2302.03169  (2023)

\bibitem{yang2021launchpad}
Yang, F., Barth-Maron, G., Stańczyk, P., Hoffman, M., Liu, S., Kroiss, M.,
  Pope, A., Rrustemi, A.: Launchpad: A programming model for distributed
  machine learning research. arXiv preprint arXiv:2106.04516  (2021),
  \url{https://arxiv.org/abs/2106.04516}

\bibitem{yu2022coca}
Yu, J., Wang, Z., Vasudevan, V., Yeung, L., Seyedhosseini, M., Wu, Y.: {CoCa:
  Contrastive captioners are image-text foundation models}. In: Transactions on
  Machine Learning Research (2022)

\bibitem{zhai2022scaling}
Zhai, X., Kolesnikov, A., Houlsby, N., Beyer, L.: Scaling vision transformers.
  In: Proceedings of the IEEE/CVF Conference on Computer Vision and Pattern
  Recognition. pp. 12104--12113 (2022)

\bibitem{zhai2023sigmoid}
Zhai, X., Mustafa, B., Kolesnikov, A., Beyer, L.: Sigmoid loss for language
  image pre-training. arXiv preprint arXiv:2303.15343  (2023)

\end{thebibliography}
